%% file: root.tex
\documentclass[letterpaper, 10 pt, conference]{ieeeconf}  % Comment this line out if you need a4paper

\IEEEoverridecommandlockouts                              % This command is only needed if 
\usepackage{graphicx} % for pdf, bitmapped graphics files
\usepackage{tabularx}
\usepackage{multirow}
\usepackage{hyperref}

\title{\LARGE \bf``Don't forget to put the milk back!" \\Dataset for Enabling Embodied Agents to Detect Anomalous Situations
}

\author{James F. Mullen Jr,$^{1,2}$ Prasoon Goyal,$^{1}$ Robinson Piramuthu,$^{1}$ Michael Johnston,$^{1}$ \\ Dinesh Manocha$^{2}$, and Reza Ghanadan$^{1}$% <-this % stops a space
\thanks{$^{1}$Amazon Science
        {\tt\small \{ prasog,robinpir,mjohnstn,   ghanadan\} @amazon.com}}%
\thanks{$^{2}$University of Maryland
        {\tt\small mullenj@umd.edu, dmanocha@umd.edu}}%
}

\begin{document}

\maketitle
\thispagestyle{empty}
\pagestyle{empty}

%%%%%%%%%%%%%%%%%%%%%%%%%%%%%%%%%%%%%%%%%%%%%%%%%%%%%%%%%%%%%%%%%%%%%%%%%%%%%%%%
\begin{abstract}

Home robots intend to make their users lives easier. Our work assists in this goal by enabling robots to inform their users of dangerous or unsanitary anomalies in their home. Some examples of these anomalies include the user leaving their milk out, forgetting to turn off the stove, or leaving poison accessible to children. To move towards enabling home robots with these abilities, we have created a new dataset, which we call SafetyDetect. The SafetyDetect dataset consists of 1000 anomalous home scenes, each of which contains unsafe or unsanitary situations for an agent to detect. Our approach utilizes large language models (LLMs) alongside both a graph representation of the scene and the relationships between the objects in the scene. Our key insight is that this connected scene graph and the object relationships it encodes enables the LLM to better reason about the scene --- especially as it relates to detecting dangerous or unsanitary situations. Our most promising approach utilizes GPT-4 and pursues a categorization technique where object relations from the scene graph are classified as normal, dangerous, unsanitary, or dangerous for children. This method is able to correctly identify over 90\% of anomalous scenarios in the SafetyDetect Dataset. Additionally, we conduct real world experiments on a ClearPath TurtleBot where we generate a scene graph from visuals of the real world scene, and run our approach with no modification. This setup resulted in little performance loss. The SafetyDetect Dataset and code will be released to the public upon this papers publication.

\end{abstract}
% Needs to be 6 pages with figures
%%%%%%%%%%%%%%%%%%%%%%%%%%%%%%%%%%%%%%%%%%%%%%%%%%%%%%%%%%%%%%%%%%%%%%%%%%%%%%%%

\input{Sections/1-Intro}
\input{Sections/2-Related}
\input{Sections/3-Dataset}
\input{Sections/4-Baselines}
\input{Sections/5-Experiments}
\input{Sections/6-Conclusion}

%%%%%%%%%%%%%%%%%%%%%%%%%%%%%%%%%%%%%%%%%%%%%%%%%%%%%%%%%%%%%%%%%%%%%%%%%%%%%%%%

{\small
\bibliographystyle{IEEEtran}
\bibliography{refs}
}

\end{document}

%% file: Sections/1-Intro.tex
\section{Introduction}
Detecting anomalies consisting of unsafe and unsanitary conditions in the home is key functionality required for home robots to be useful for users. For instance, if you forget to put your milk back in your fridge, or leave the front door ajar, it would be great if a home robot notified you, or solved the problem itself. Additionally, users with children will materially benefit from a robot that can monitor the environment for their children's safety.

 \begin{figure}[t]
     \centering
     \includegraphics[width= 0.8\linewidth]{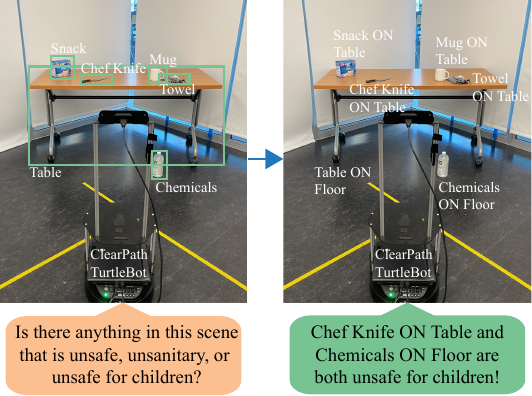}
     \vspace{-0.2cm}
     \caption{In this work we aim to enable embodied agent to detect unsafe and unsanitary conditions in the home. For this, we first create a new, unique dataset with unsafe and unsanitary conditions to detect. We then hypothesize that Large Language Models (LLMs) contain the knowledge needed to logically operate on these conditions. Our approach creates LLM prompts that leverage object relationships in the scene from a scene graph (like those in the right image with objects being nodes and relationships being edges), and classifies them. In addition to testing on our dataset, we tested in the real world using the ClearPath TurtleBot in scenarios like that shown here.
     }
     \label{fig:coverimg}
     \vspace{-0.5cm}
 \end{figure}

These types of scenarios can present a real danger to people in the home. For example, 31\% of home cooking fires are caused by unattended equipment \cite{fire}, over 42,000 people died from falls sustained at home or at work \cite{falls}, and accidents, including poisoning and suffocation, are the leading cause of death for children in the United States \cite{children}. If a home robot can monitor the stove to make sure it is properly turned off, police the environment for tripping hazards, and monitor the home for accessible poisons or suffocation hazards, many of these fires, injuries, or deaths can be prevented. 

Building embodied agents that can perform household tasks has seen a lot of recent interest from the robotics and embodied AI communities. Some commonly approached problems include navigation \cite{safenav, robotrust}, instruction following, and embodied question answering \cite{jesse1, cvdn, dialfred}. Each of these tasks defines a precise goal, e.g. navigating to a set location, moving objects to a set location, or answering a question correctly. However, our use case, detecting anomalies in the home, is a poorly formulated problem consisting of an incredibly diverse and disparate set of examples. As such, it is difficult to extensively hard-code into a robots behavior. One would have to record a rule for where every object can or cannot be kept, and what state it must be in. Additionally, these rules would have to adapt to an unknown and unique set of objects and personal preferences in a household. We hypothesize that recent advances in Large Language Models (LLMs) \cite{gpt3, openai2023gpt4, touvron2023llama} could enable our use case by providing diverse knowledge on the home that can be leveraged to detect potentially unsafe or unsanitary conditions. However, there are no sources of data that contain environments with these conditions to test potential methods on. Additionally, we show in this work that previous approaches to prompting LLMs are unable to effectively extract the knowledge necessary to detect unsafe and unsanitary conditions.

\textbf{Main Contributions:} We introduce the SafetyDetect dataset to benchmark the ability of embodied AI agents to infer unsafe or unsanitary situations in the house. \autoref{fig:coverimg} showcases our task where an agent is spawned randomly in an unknown environment, with an unknown set of potentially unsafe or unsanitary conditions. Without explicit instructions, the agent must discover any potential anomalies and report them to the user.

The SafetyDetect dataset is challenging due to the diverse and unrelated nature of the possible anomalies in the scene. For example, medicine on the ground (poison hazard for children) is hardly connected to moldy produce on the counter (sanitation hazard). Additionally, agents are not trained on a base environment allowing them to simply detect changes. Agents are required to use some form of knowledge to deduce the danger to the user. 

We propose baseline solutions that encapsulate knowledge leveraged from LLMs and demonstrate that this serves as an effective solution for the SafetyDetect dataset. Our key finding is that utilizing a scene graph is the most effective way of informing the LLMs of the proper scene context and semantic information for scene understanding and spatial reasoning. Baseline methods that do not use a scene graph perform very poorly on the SafetyDetect dataset.

\begin{table*}[t] 
    \centering
    \begin{tabular}{l|c c c c c c} 
        Dataset & Goal & Scenarios & Object Categories & Object Models & Scenes\\ [0.5ex] 
        \hline\hline
        Transport Challenge & Geometric & Inf & 50 & 112 & 15\\
        \hline
        Behavior & Predicate & 1000 & 391 & 1217 & 15 \\
        \hline
        My House, My Rules & Human Preferences & 75 & 12 & 12 & 2 \\
        \hline
        Housekeep & Human Preferences & 585 & 268 & 1799 & 14\\
        \hline
        \textbf{Ours} & \textbf{Anomalies} & 1000 & 192 & 1163 & 7\\
        \hline
    \end{tabular}
    \vspace{5.0px}
    \caption{Comparison of SafetyDetect to other similar benchmarks. SafetyDetect is comparable to the other datasets in scale while approaching a different end task. Note that we can use procedural generation to go beyond 1000 scenarios and 7 scenes.}
    \label{tab:dataset}
    \vspace{-0.8cm}
\end{table*}
%\textbf{Ours} & \textbf{Anomalies} & 1000 & 16+176 & 25+1138 & 7+\\

Our main contributions are as follows:
\begin{enumerate}
    \item We present the SafetyDetect dataset, a new, first-of-its-kind dataset, built off the VirtualHome simulator\cite{puig2018virtualhome}, aimed at enabling researchers to create embodied agents that can detect unsafe or unsanitary conditions in the home. This dataset, at release, contains 1000 scenarios for users to explore and solve. We additionally provide information about each scenario which informs user preference and how the robot should report a given anomaly to the user.
    \item We present an LLM-based method that leverages a scene graph, categorization, and chain of thought prompting to perform exceptionally well on a simplified version of this task, with an anomaly detection rate of over 95\%.
    \item We show that use of the scene graph when creating the LLM prompt is important for performance on our SafetyDetect dataset through its ability to provide scene information to the LLM in a concise way.
    \item We explore the sim-to-real transfer of our method tested on the SafetyDetect dataset and demonstrate that the SafetyDetect dataset is an adequate representation of real world scenarios. We do this by running our method on a ClearPath TurtleBot and having it detect unsafe or unsanitary anomalies in a real-world home environment.
\end{enumerate}

%% file: Sections/2-Related.tex
\section{Related Work}
\subsection{Similar Datasets or Benchmarks}
While to our knowledge no previous work has approached household anomaly detection as a task, there are some alternate tasks that are similar in scope or implementation. Behavior1K \cite{li2023behavior1k} is a simulation benchmark where the robots must complete 1000 everyday tasks. \cite{kant2022housekeep} and \cite{wu2023tidybot} both focus on the idea of cleaning up clutter in the home and placing things in their proper locations. \cite{kant2022housekeep} is the most similar to our work as in both tasks, agents are placed randomly into the environment and must find the issues, unsafe situations in our case and misplaced objects in theirs. Unique to our work is the specific use case of the detection of unsafe or unsanitary conditions. Approaches on the HouseKeep dataset are built solely around their task and would need significant modification to work on SafetyDetect.

\subsection{Language-Based Embodied AI}
Using language to inform robotic agents is a popular task in literature, with work including using generalized grounding graphs \cite{tellex2} for robot manipulation \cite{ggs, gg2} to performing language-guided navigation \cite{safenav, robotrust}.
%  in autonomous driving \cite{langdynamic}, and also in drone-control \cite{dronelang1}
Tellex et al. \cite{tellex1} recently presented a useful survey on using language from a robotics perspective.

More recent work tackling this problem by Thomason et al. \cite{jesse1, cvdn} and Gao et al. \cite{dialfred} has explored the use of human-robot dialogue to gather relevant information for completing tasks. Different from these works, we are focused on using natural language derived from a scene graph as a medium for scene understanding. However, parsing and utilizing natural language is very relevant in our work, and we are motivated by the techniques developed by these papers.

With the advent of ChatGPT \cite{gpt3}, LLaMA \cite{touvron2023llama}, FLAN-T5 \cite{weiflant5}, and other LLMs, the field of Embodied AI worked to leverage them to improve performance on their tasks. Dorbala et al. \cite{dorbala} use language models to inform navigation for object goal navigation. Singh et al. \cite{progprompt} use language models to write code that solves a given task. We differ from these methods in terms of task and prompting technique. Specifically, creating the SafetyDetect dataset to explore the detection of unsafe conditions is substantially different from the prior work which is generally focused on question-answering \cite{driess2023palme, huang2022inner}, task-completion \cite{progprompt, saycan2022arxiv}, or object goal navigation \cite{dorbala}. Additionally, our utilization of the scene graph to provide context to the LLM when prompting is a key differentiator from these methods.

\subsection{Using the Scene Graph}
Scene graphs are a common method of representing a scene in computer graphics and 3D modeling where, generally, nodes of the graph are objects and edges are relationships. Additionally, creating a scene graph from images is a popular problem in the computer vision community \cite{yang2018graph, xu2017scene}. Many simulation platforms for embodied AI are built on top of a scene graph including  both Habitat \cite{savvaHabitatPlatformEmbodied2019} and VirtualHome \cite{puig2018virtualhome}. This makes a scene graph a relatively easy to access source of information, both in simulation and real world environments, for methods trying to solve embodied AI tasks. However, few works have attempted to leverage the scene graph for scene understanding with LLMs. 

In contrast to our approach with \textit{scene} graphs, existing LLM literature uses \textit{knowledge} graphs as a means of finding information to the LLM as context. These methods generally conduct a semantic search of the knowledge graph to find said information \cite{knowledgegraph, knowledge2, zhang2023graph}.
The closest work to ours is SayPlan \cite{rana2023sayplan} which in fact uses a scene graph, but only uses it as a knowledge graph and a means of conducting a semantic search on the scene to provide relevant information to the LLM based on the task at hand. In contrast to this, we utilize the scene graph to create strings that encapsulate object relationships to then feed into the LLM alongside a classification approach that does not allow for filtering the graph extensively.

%% file: Sections/3-Dataset.tex
\section{Adding Anomalies to the Home: The SafetyDetect Dataset}
\subsection{Creating the SafetyDetect Dataset}
\textbf{Task Definition:} In SafetyDetect, an embodied agent is tasked with finding any unsafe or unsanitary conditions in the home, and reporting them to the user. Additionally, the SafetyDetect dataset requires the agent to only report conditions that meet the users preferences. For example, if there are no children in the home, the agent should not report a situation that is only dangerous for children. The agent may also be provided a graph representation of the scene, similar to that which can be created by \cite{yang2018graph} in the real world, with nodes of the graph being the objects that make up the scene, and edges of the graph being relationships like `ON,' `INSIDE,' and `FACING.'

\begin{figure*}[t]
     \centering
     \includegraphics[width= \linewidth]{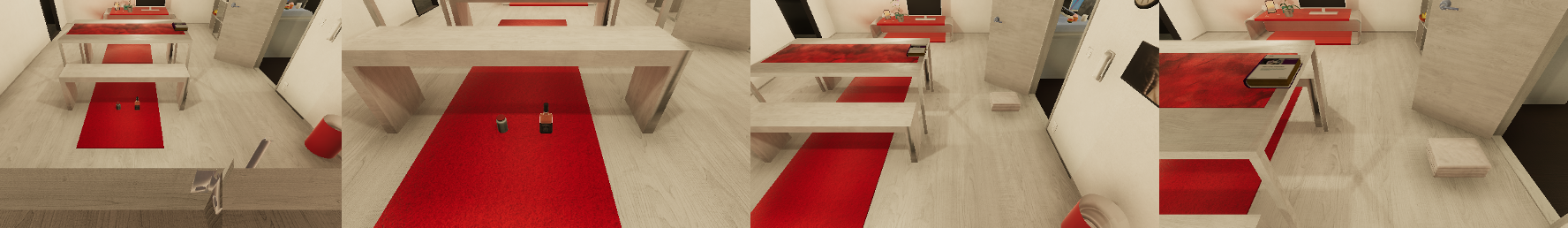}
     \vspace{-0.75cm}
     \caption{A sampling of images from the SafetyDetect dataset showing unsafe conditions. In one of the images, medication and alcohol are on the floor and dangerous for children. In another, a pile of clothes are in the doorway - a tripping hazard for users.
     }
     \label{fig:qual}
     \vspace{-0.6cm}
 \end{figure*}

\textbf{Simulator and Scenes:} We use VirtualHome \cite{puig2018virtualhome} as the basis for the SafetyDetect dataset. VirtualHome was chosen for its ease of use, existing support community, and relative simplicity of adding novel objects. VirtualHome contains seven pre-built scenes, but supports procedural generation to create an unlimited number of valid home environments. For the SafetyDetect dataset we rely on the pre-built scenes, but we release code to allow users to procedurally generate new scenes and data samples with unlimited diversity.

\begin{table}[t] 
    \centering
    \begin{tabular}{l c c} 
        Category &  Class Name & Number of Occurances\\ [0.5ex] 
        \hline\hline
        \multirow{7}{4em}{Safety} & Spills & 195\\
        & Tripping Hazard & 202\\
        & Broken Items & 214\\
        & Candle On & 190\\
        & Front Door Open & 181\\
        & Stove On & 207\\
        \hline\hline
        \multirow{3}{4em}{Sanitation} & Refrigerated/Frozen foods out & 210\\
        & Expired Produce & 199\\
        & Fridge/Freezer Open & 184\\
        \hline\hline
        \multirow{5}{4em}{Safety for Children}& Choking Hazard & 201\\
        & Sharp Objects & 201\\
        & Poison: Cleaning Products & 192\\
        & \multirow{2}{12em}{\centering Poison: Medication \& Beauty Products} & \multirow{2}{1.5em}{198}\\
    \end{tabular}
    \vspace{10.0px}
    \caption{Outline of all of the classes of anomalies in the dataset as well as the number of times they occur in the dataset.}
    \label{tab:dataset-breakdown}
    \vspace{-1.0cm}
\end{table}

\textbf{Anomalies and Objects:} To create SafetyDetect, we first outlined our target set of hazards that fit the overarching categories of unsafe conditions, unsanitary conditions, and conditions which are dangerous for children. The set of hazards, which are the classes of the dataset, is outlined in \autoref{tab:dataset-breakdown}, alongside the number of occurrences of each in the dataset. This was created primarily by referencing large scale statistical studies of major household dangers for users or their children. For example, \cite{children} outlines the major causes of death for children in the United States. After filtering out dangers to children not relevant to the household environment (auto accidents), and those we did not feel comfortable addressing (firearm related incidents), we chose to cover the most common remaining examples including choking, poison hazards, and sharp objects. Additional individual examples were created through a small scale user study soliciting examples from prospective users. This added classes like `broken items' which adds the detection of items like shattered glasses or mugs that may have been broken by pets.

To embody and visualize these hazard classes in the simulator, we had to determine what objects were necessary to recreate each hazard \textit{and} where those objects could be placed to constitute the hazard while remaining logically plausible. We began by manually noting obvious examples. For example, a candle flame hazard would require a candle object with an active flame, and a tripping hazard must be on the floor. For each hazard class, we expanded our samples by prompting an LLM to first produce sample objects, and then for sample placement location examples that would create the target hazard using the other objects present in the scene. The LLM's output was manually filtered to examples that made logical sense and could be produced in the simulator. 

Many objects we required were not present in the VirtualHome simulator. To address this, we added relevant objects from Google Scanned Objects \cite{downs2022googlescannedobjects}, ReplicaCAD \cite{savvaHabitatPlatformEmbodied2019}, Fantastic Breaks \cite{lamb2023fantastic}, and YCB Objects \cite{ycb}. We additionally created new objects and modified similar objects using Blender to cover cases where suitable objects could not be found. For example, we created a spill object and texture, and re-textured various fruits and vegetables to appear as their expired or rotten counterparts. In total, we add over 30 new objects to VirtualHome while leveraging many objects native to the simulator. The final assortment of objects and placement locations results in 967 unique anomalies split among the 13 hazard classes. Through combining these anomalies, countless unique scenes can be created.

\textbf{Generating the Dataset:}
To generate the final dataset, we randomly sampled 1000 i.i.d scenarios, each of which utilizes one prebuilt scene, contains 0-5 hazards, and shares a set of user preferences. Each hazard consists of the hazard class, a selected object affiliated with said class, and a valid placement of that object to embody the hazard. The final scene graph after placing the object into the scene in VirtualHome is provided. Future users can visualize our scenarios, add more samples with the prebuilt environments, or procedurally generate new scenes and scenarios.

\textbf{Agent:} VirtualHome contains an agent for use in exploring the scene. The agent is embodied as a human but contains a RGB camera who's location on the agent and Field of View (FOV) we can alter. This agent can navigate around the scene using the VirtualHome API. This includes commands for walking towards a room or object, forward in the direction of travel, and turning a certain angle.

\subsection{Using the SafetyDetect Dataset}
The central challenge of the SafetyDetect dataset is understanding how to detect unsafe or unsanitary conditions in the home and notify the user of their existence. Our goal is to capture a comprehensive but not necessarily all-inclusive set of these conditions so researchers can test their methods before deploying them into a consumer environment. These scenarios were collected through surveys of prospective users and expert documentation. This allowed us to produce a diverse set of scenarios for exploration.

\textbf{Episodes:} Each episode in the SafetyDetect dataset instantiates one of the seven base scenes of VirtualHome before extending it with 0-5 unsafe or unsanitary conditions. The agent itself is spawned randomly into the environment with no prior knowledge of it and can explore the environment to find potentially unsafe or unsanitary conditions. We call this set of conditions $\mathcal{A}$, where each individual anomaly is denoted with $a_i$. An agent is then placed randomly into the scene. Next, we define for the agent the relevant context of the scene, specifically the presence of children in the house. The researcher must then create a strategy for the agent that allows it to detect the anomalies in the home, and reproduce $\mathcal{A}$. Our solution is described in the next section.

\textbf{Evaluations:} We evaluate agents/detection schemes for effectiveness and efficiency. All metrics are reported per episode and are then aggregated across multiple episodes to report the averages and standard errors. These metrics are utilized in similar literature including \cite{kant2022housekeep}. 

\begin{enumerate}
    \item \textbf{Anomaly Success (AS)}: Fraction of all anomalies found in a given episode. This is essentially the true positive detection rate.
    \item \textbf{Conditioned Anomally Success (CAS)}: Fraction of anomalies reported correctly minus those reported incorrectly depending on the given context. For example, in a home without children, a knife being left on the counter is not a problem and should not be reported. Reporting in a situation like this could be annoying to the user, hampering the agent's effectiveness.
\end{enumerate}

Additionally, we track the true positive rate for each anomaly individually. Tracking of false positive and false negative results is also conducted automatically but analysis must be conducted manually. Code released with the SafetyDetect dataset allows users to easily glean this information for their own methods.

%% file: Sections/4-Baselines.tex
\section{Finding Anomalies in the Home}

For our proposed method we employ the GPT-4 Large Language Model (LLM) \cite{openai2023gpt4} to apply knowledge about what scenarios may be unsafe, unsanitary, or dangerous for children in a scene. Specifically, we hypothesize that LLMs contain domain knowledge about what is safe in a home environment, and what might be dangerous, unsanitary, or dangerous for children. To enable performance from the LLM, we provide two key insights: 1) utilizing a scene graph provides the context needed for scene understanding and reasoning, and 2) classification is an effective approach for detecting specific situations in a scene.

\begin{figure*}[t]
     \centering
     \includegraphics[width= \linewidth]{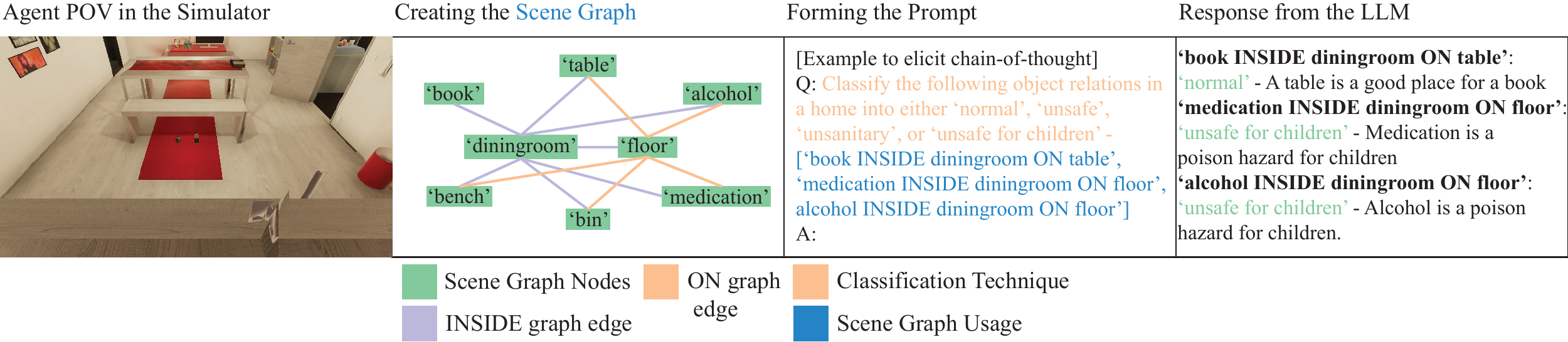}
     \vspace{-0.75cm}
     \caption{The flow of our method is depicted here. We first get the scene graph before using it to formulate a prompt with asks the LLM to categorize the object relations. The model must then use commonsense reasoning to categorize these object relations effectively.
     }
     \label{fig:method}
     \vspace{-0.6cm}
 \end{figure*}

\textbf{Using the Scene Graph.} For the scene graph, we assume that other methods exist, or can be created, to generate a scene graph that represents object relations in the scene. Some examples of these methods include \cite{xu2017scene}, and \cite{yang2018graph}. The VirtualHome \cite{puig2018virtualhome} simulator provides a scene graph of this type natively. In the scene graph, each object is a node and the relation is an edge. For example, an apple sitting on a kitchen table would have its initial node with a class of apple, and then two edges, on showing it `ON' the kitchen table, and another showing it `INSIDE' the kitchen. The use of the scene graph is imperative to our method and what sets it apart from many other approaches which generally provide either a textual description of the environment, or a listing of classes found in the immediate area \cite{dorbala}. 

For our LLM prompts, we specifically parsed the scene graph for the edge representing the room the object was inside, and the edge for what the object was on top of. This produced object relation strings like `apple INSIDE kitchen ON counter' or `medication INSIDE livingroom ON floor.' These strings were in turn fed directly into GPT-4 alongside the rest of the prompt outlining the classification problem and providing some examples to elicit chain-of-thought.

\textbf{Classifying the Scene Graph.} The second key aspect of our proposed method is classifying relationships in the scene graph as a way of differentiating normal and dangerous situations. Specifically, our prompt for GPT-4 asks it to classify object relations as `normal', `unsafe', `unsanitary', or `unsafe for children'. This classification approach improved performance as similar methods that asked the LLM to directly detect anomalous object relationships would frequently produce false negatives.

\textbf{Building the Prompt.} We also provide the LLM with a series of example anomalies given as an object relation pulled from the scene graph, followed by the correct classification and an explanation. We employ the popular chain-of-thought prompting technique to provide these examples. The prompt we utilize for our method is below, where [CONTENT] is replaced by the object relations for the LLM to classify:
\begin{quote}
    \begin{em}
    \textbf{Q:} Classify the following object relations in a home into either `normal', `unsafe', `unsanitary', or `unsafe for children' - [`medication INSIDE livingroom ON floor', `rawmeat INSIDE kitchen ON counter']
    
    \textbf{A:} `medication INSIDE livingroom ON floor': `unsafe for children' as medication is a poison hazard
        
        `rawmeat INSIDE kitchen ON counter': `unsanitary' as rawmeat requires refrigeration and will go bad if left unrefrigerated.
    
    \textbf{Q:} Classify the following object relations in a home into either `normal', `unsafe', `unsanitary', or `unsafe for children’ – [CONTENT]
    
    \textbf{A:}\end{em}
\end{quote}

\textbf{Parsing and Evaluating the Response.} To parse through the verbose responses, we again prompt a LLM, this time GPT-3.5 Turbo for cost savings with minimal performance loss. We ask the model to simplify the previous response into a one word classification fitting the categories above. For example, a response including ```medication INSIDE livingroom ON floor': `unsafe for children' as it presents a possible poison hazard,'' would be filtered into just `unsafe for children.' This allows us to evaluate the models performance effectively despite the verbose responses provided through chain-of-thought prompting. This parsing method injected a small amount of error into the system, with around 1-2\% of responses being improperly parsed in our method.

%% file: Sections/5-Experiments.tex
\section{Experiments and Results}
\subsection{Baseline Results}
\begin{table}[t] 
    \centering
    \begin{tabular}{l|c c} 
        Method & AS $\uparrow$ & CAS $\uparrow$\\ [0.5ex] 
        \hline\hline
        Object List + E, C & 12.4 & 5.1 \\
        \hline
        Scene Description + E, C & 21.5 & 13.6 \\
        \hline
        \hline\hline
        SG, no C & 18.8 & -\\
        \hline
        E + SG, no C  & 38.8 & -\\
        \hline\hline
        SG \& C & 83.2 & 80.8\\
        \hline
        E + SG \& C & 94.0 & 88.8\\
        \hline
        \textbf{CoT + SG \& C (Ours)} & \textbf{96.0} & \textbf{90.5}\\
        \hline
    \end{tabular}
    \vspace{5.0px}
    \caption{Comparison of performance on SafetyDetect. E represents the inclusion of example, SG represents the use of the Scene Graph, C represents the inclusion of categorization. Note that the methods that utilize the scene graph and categorization are significantly more competitive than those that do not.}
    \label{tab:results}
    \vspace{-0.7cm}
\end{table}

\begin{table}[t] 
    \centering
    \begin{tabular}{l|c c} 
        Method & AS $\uparrow$ & CAS $\uparrow$\\ [0.5ex] 
        \hline\hline
        % LLaMA-7B & 12.3 & 4.1\\
        % \hline
        FLAN-T5-Large & 81.3 & 51.2\\
        \hline
        GPT-3.5-Turbo & 94.0 & 88.8\\
        \hline
        \textbf{GPT-4 (Ours)} & \textbf{96.0} & \textbf{90.5}\\
        \hline
    \end{tabular}
    \vspace{5.0px}
    \caption{Comparison of performance on SafetyDetect for different Large Language Models. LLaMA is not included as we could not get it to work well on this task.}
    \label{tab:llm-ablate}
    \vspace{-0.9cm}
\end{table}

\textbf{Use of The Scene Graph and Classification.} We first tested against the use of the scene graph and classification techniques in the prompts. We find that when providing scene information to the LLMs, standard techniques like using object lists or a scene descriptions had difficulty understanding when an anomaly was present. For example, when providing a simple object list, its difficult for the model to know whether an object is in a safe space or not. As such, it tended to err on the side of reporting normalcy. It did relatively well with specific cases like spills or rotten fruit as the class name was sufficient to detect an anomaly. The scene description-based prompts performed better, with a 21.5\% detection rate, but it still struggled to pick out dangerous and unsanitary conditions. It performed similarly to the object list method for objects whose class names alone were enough to detect the anomaly and added some performance in other cases like sharp items and refrigeration required items.

When utilizing the scene graph as the basis for communicating the scene context, performance nearly doubles when continuing to provide examples but removing classification. When adding classification back in, performance doubles again to a nearly 4.5$\times$ improvement over the scene description-based method for a 94\% detection rate. Reformatting the prompt to fit a chain-of-thought scheme further boosted the performance slightly to a 96\% detection rate.

Upon further inspection, the logic provided by GPT-4 in response to our method was consistent with real-world commonsense. Anomalies were not only correctly classified, but the logic behind their classification matched that in expert sources and from our sample users. %One interesting behavior was the method consistently calling `spills' unsanitary \cite{and} unsafe as a spill can encourage bacteria and mold growth as well as be a slipping hazard.  
In fact, it found anomalies built into the scene graph that we had not placed there, but were potential hazards. Some scenes were built in VirtualHome through stacking objects on top of each other to create a visually new object, but the scene graph retained this stacked structure and produced illogical relationships. Our method would consistently pick out these scenarios as anomalous. As the underlying structure of the simulator had many of these anomalies, it became difficult to track a false positive rate. Anecdotally, we saw very few false positives in the our method's responses.

\textbf{Ablation on Models.} We also test against the specific use of GPT-4 by deploying the same Chain-of-Thought + Scene Graph \& Classification prompt on GPT-3.5-Turbo, FLAN-T5-Large, and LLaMA-2. We find that GPT-3.5 produces a similar level of performance to GPT-4 with FLAN-T5-Large also producing respectable detection rates above 80\%. We had difficulty applying LLaMA to this task as it frequently hallucinated new object relations that we did not provide. This made parsing the response increasingly difficult and severely impacted performance. Further experiments with different prompting approachesay be able to improve upon these LLaMA results.

\subsection{Real World Experimentation}
The goal of our dataset is to enable new use cases in home robotics. As such, it is important to validate the use of the scene graph in the real world and explore how best to create the scene graph for our purposes. For this experiment, we set up a ClearPath TurtleBot to navigate the local environment while running a slightly modified version of \cite{yang2018graph} that builds scene graphs similar to that from VirtualHome for use alongside our CoT + SG \& C method.

Our real world scene consisted of one room with a small number of objects, including 1-3 anomalies. The TurtleBot captures images from $360^\circ$ and feeds them into our modified version of \cite{yang2018graph} that produces an adequate scene graph. We then run our algorithms verbatim on the scene graph. We conduct 20 experiments that covered a subset of eight of our anomaly classes in the SafetyDetect task. 

We find that performance is similar between the simulator and the real world. Creating the scene graph through \cite{yang2018graph} also enabled increased performance on some anomalies, such as obstructions, which were not captured well in the scene graph created by the VirtualHome simulator. Some anomalies suffered worse performance as the scene graph generator we utilized was unable to distinguish between good and rotten produce, and often did not detect spills. We believe that many of these issues are easily solved by retraining the object detection components of the scene graph builder. Overall, our detection rate dropped to 82.7\%. Through this experiment, we believe that if provided with a ground truth scene graph, there is little reason why our LLM-based approach would not be able to sustain our simulator-based detection rates of over 95\% with few false positives.

\begin{figure}[t]
     \centering
     \includegraphics[width= 0.7\linewidth]{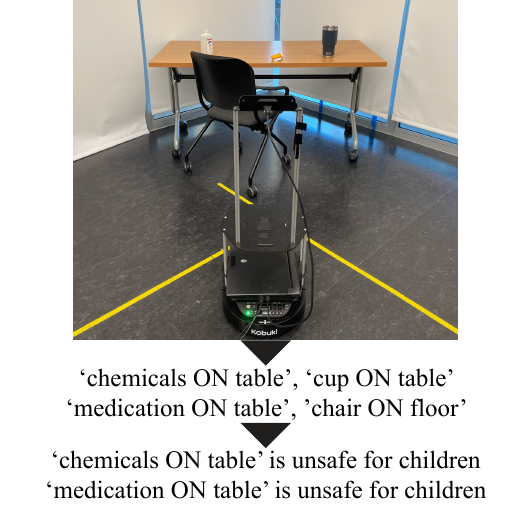}
     \vspace{-0.3cm}
     \caption{We deploy a ClearPath TurtleBot in the real world. We use existing methods to generate an effective scene graph before utilizing our method to detect anomalies in the scene. In this example, medication and cleaning products are on the table. This is captured in the scene graph and detected by the model as unsafe for children.
     }
     \label{fig:real}
     \vspace{-0.6cm}
\end{figure}

%% file: Sections/6-Conclusion.tex
\section{Conclusions, Limitations, and Future Work}

Home robots intend to make their users lives easier. One way they may do this is by informing users of issues or anomalies in the home. In this work, we moved towards enabling home robots with these abilities by creating a new dataset, built on top of the popular VirtualHome platform, that contains 1000 dangerous or unsanitary scenarios for an agent to detect. We also propose an LLM-based approach that utilizes a scene graph and classification approach. This methods identifies over 90\% of anomalous scenarios in our dataset. Additionally, we show that these techniques remain viable in the real world.

\textbf{Limitations.} The methods we proposed are limited by perception, specifically how the scene graph is created. For example, in the scene graph created by the VirtualHome simulator, there was nothing indicating obstructions in the scene. To illustrate this, a relation of `box INSIDE livingroom ON floor' is okay, but physically the box may be in the doorway creating a tripping hazard. While we assumed perfect perception in creating the scene graph in the simulator, alternate approaches without this assumption would produce weaker results.

\textbf{Future Work.} Continuation work with multi-modal LLMs like GPT-V could be valuable to both address the perception perception and solve the problem in one shot. The next logical step in this work is to get the agent to preemptively solve the anomaly by adding on a task planning and completion task. For the real world deployment of this work, significant effort needs to be put into optimizing prompts and altering the scheme to lower the costs of querying the LLM. 
%Also valuable would be a further exploration of what additional knowledge about the scene could be encoded into a scene graph to allow for even better detection results.
%We would also like to pursuing approaches that leverage multi-modal LLMs and try to exploit the entire visual of the scene instead of a compressed form like the scene graph would be interesting.